\documentclass[a4paper]{article}
\usepackage[margin=0.5in]{geometry}
\usepackage{times}
\usepackage{epsfig}
\usepackage{graphicx}
\usepackage{amsmath}
\usepackage{amssymb}
\usepackage{multirow}
\usepackage[ruled,vlined]{algorithm2e}

\usepackage{mypackages}
\pdfoutput=1
\hypersetup{linkcolor=red, urlcolor=magenta, citecolor=green, colorlinks=true}

\usepackage{epsfig}
\usepackage{graphicx}
\usepackage{amsmath,amssymb} 
\usepackage{color}

\definecolor{coltabhead}{RGB}{255,255,255}
\definecolor{coltabheadtext}{RGB}{0,0,0}

\def\myalgoname{UCapsNet}
\def\primarycapsdown{\texttt{PCD}}
\def\primarycapsup{\texttt{PCU}}
\def\doubleblockdown{\texttt{DBD}}
\def\doubleblockup{\texttt{DBU}}

\def\mybn{\texttt{BN}}
\def\myrelu{\texttt{ReLU}}
\def\myconv{\texttt{Conv}}
\def\mymaxpool{\texttt{MaxPool}}
\def\myTconv{\texttt{TransposeConv}}
\def\myflatten{\texttt{Flatten}}
\def\myupsample{\texttt{UpSample}}

\begin{document}

\title{Collaboration among Image and Object Level Features for Image Colourisation}

\author{Rita Pucci\\
University of Udine\\
Udine, Italy\\
{\tt\small rita.pucci@uniud.it}
\and
Christian Micheloni\\
University of Udine\\
Udine, Italy\\
{\tt\small christian.micheloni@uniud.it}

\and
Niki Martinel\\
University of Udine\\
Udine, Italy\\
{\tt\small niki.martinel@uniud.it}
}

\twocolumn[{%
\renewcommand\twocolumn[1][]{#1}%
\maketitle
\begin{center}
    \vspace{-2.1em}
	\centering
    \includegraphics[width=\linewidth,height=0.24\textheight]{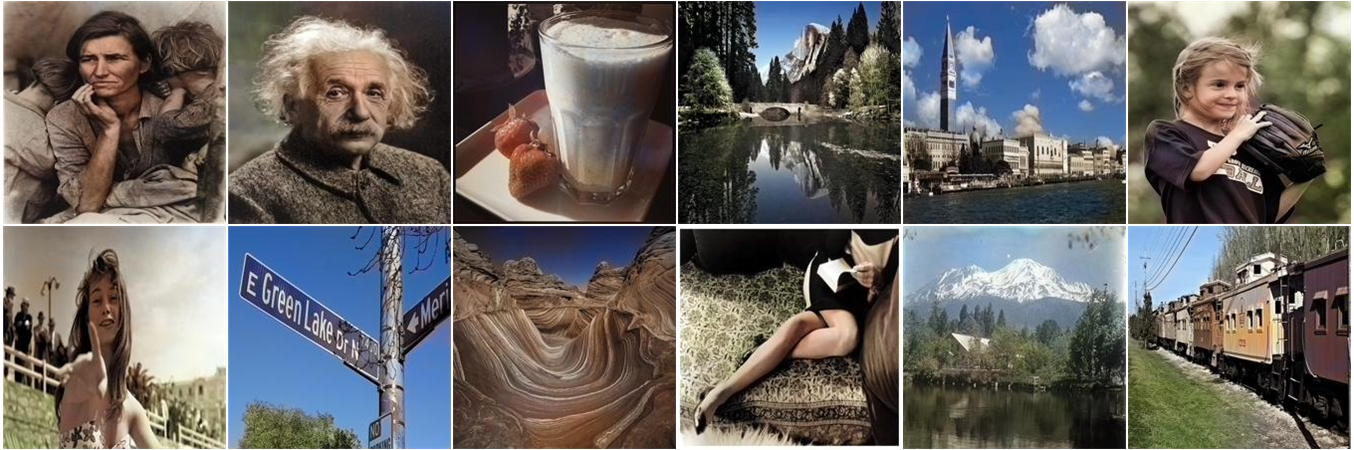}
	\captionof{figure}{Achieved automatic colourisation results. We introduce a self-supervised colourisation model that disentangles image (context) and object features to produce plausible and colourful results on different scenes with no human intervention.}
	\label{fig:intro}
\end{center}%
}]

\begin{abstract}
\vspace{-1em}
Image colourisation is an ill-posed problem, with multiple correct solutions which depend on the context and object instances present in the input datum.
Previous approaches attacked the problem either by requiring intense user interactions or by 
exploiting the ability of convolutional neural networks (CNNs) in learning image level (context) features.
However, obtaining human hints is not always feasible and CNNs alone are not able to learn object level semantics, unless multiple models pretrained with supervision are considered.
In this work, we propose a single network, named \myalgoname, that disentangles image level features obtained through convolutions and object-level features captured by means of capsules.
Then, by  skip connections over different layers, we enforce collaboration between such disentangling factors to produce a high quality and plausible image colourisation.
We pose the problem as a classification task that can be addressed by a fully self-supervised approach, thus requires no human effort.
Experimental results on three benchmark datasets show that our approach outperforms existing methods on standard quality metrics and achieves state of the art performances on image colourisation.
A large scale user study shows that our method is preferred over existing solutions. 

\end{abstract}

\section{Introduction}
\vspace{-0.1em}
Adding a plausible colourisation to a monochromatic (greyscale) image through an autonomous system is a challenging research topic in computer vision and pattern recognition~\cite{yatziv2006fast}. 
Automatic colourisation has significant impacts on historical image/video restoration and image compression~\cite{li2020review}, while also being a useful proxy task for learning visual representations~\cite{Owens_2018_ECCV}. The colourisation task has multiple plausible choices due to the multi-modality of appearance of entities (\eg, a car can be red, blue, green)~\cite{charpiat2008automatic}. 

Following such a consideration, previous works relied on intense user-interaction procedures to guide the colourisation through  scribbles-based~\cite{levin2004colorization,huang2005adaptive,yatziv2006fast,huang2005adaptive,luan2007natural,zhang2017real,qu2006manga} and image references-based~\cite{gupta2012image,zhao2018pixel,he2018deep,deshpande2015learning} methods. Fully automatic solutions were proposed by introducing different deep architectures~\cite{Su2020CVPR,deshpande2017learning,iizuka2016let,cheng2015deep,guadarrama2017pixcolor,he2018deep,isola2017image,zhang2016colorful,larsson2016learning,ozbulak2019image,mouzon2019joint} which either considered image-level or object-level features alone, thus neglecting the importance of the interaction between the global content and the object instances in an image. Such methods also heavily relied on pre-trained models and supervised learning tasks to learn object semantics.

On the contrary, we hypothesise that the colourisation process can be tackled through (i) a single model that (ii) captures image context and the object entities (iii) without supervision to generate a plausible colourisation\footnote{\ie, that has geometric, perceptual, and semantic photo-realism.
}.
\textit{These considerations motivate us to introduce a novel approach that performs automatic colourisation by disentangling the image context from the object instances, without the need of labelled data.} 
To capture such factors, we leverage convolutional and capsules layers~\cite{sabour2017dynamic}. 
The former allows us to identify and extract features which carry information about the image context.
Through the routing by agreement routine, the latter captures the presence and the features of object entities. 
Skip connections over different layers are used to encourage a collaboration between disentangled representations, thus allowing us to leverage image-level and object-level features to produce a plausible colourisation.
We considered the fact that any colour photo can be used as a training sample by taking the image's lightness channel as input and its colour channels as the supervisory signal. Following such an intuition, we pose the colourisation problem as a self-supervised task that learns a distribution of colours for each pixel.
By exploiting such a distribution through a trainable mapping function, we capture the fact that a same object can have different colourisations (\eg, an apple can be red, green or yellow).

Concretely, our contribution is a novel fully automatic image colourisation approach that  
\begin{itemize}
	\item learns disentangled image level and object level representations to grasp information about the image context and object entities;
	\item exploits skip connection at different network depths to enforce collaboration between the disentangled features;
	\item captures the multi-modality of the colourisation problem by learning to predict a distribution of plausible colours;
	\item sets the learning problem as a self-supervised task, allowing us to tackle the colourisation problem with neither human intervention nor pre-trained models.
\end{itemize}

We have conducted a comprehensive evaluation of our model on three large scale benchmarks datasets, namely ImageNet~\cite{russakovsky2015imagenet}, COCOStuff~\cite{caesar2018coco} and Places205~\cite{zhou2014learning}.
We also assessed the qualitative performance of the proposed approach through a large scale human-based evaluation study.
Results demonstrate that our approach outperforms existing works in terms of the image colourisation quality both using common image-quality metrics as well as with respect to the human preference.
We also show its advantages in generating plausible colourisation of different entities over works leveraging labelled data to learn object semantics.

\begin{figure*}[t]
\begin{center}
\includegraphics[width=\textwidth,height=0.4\textheight]{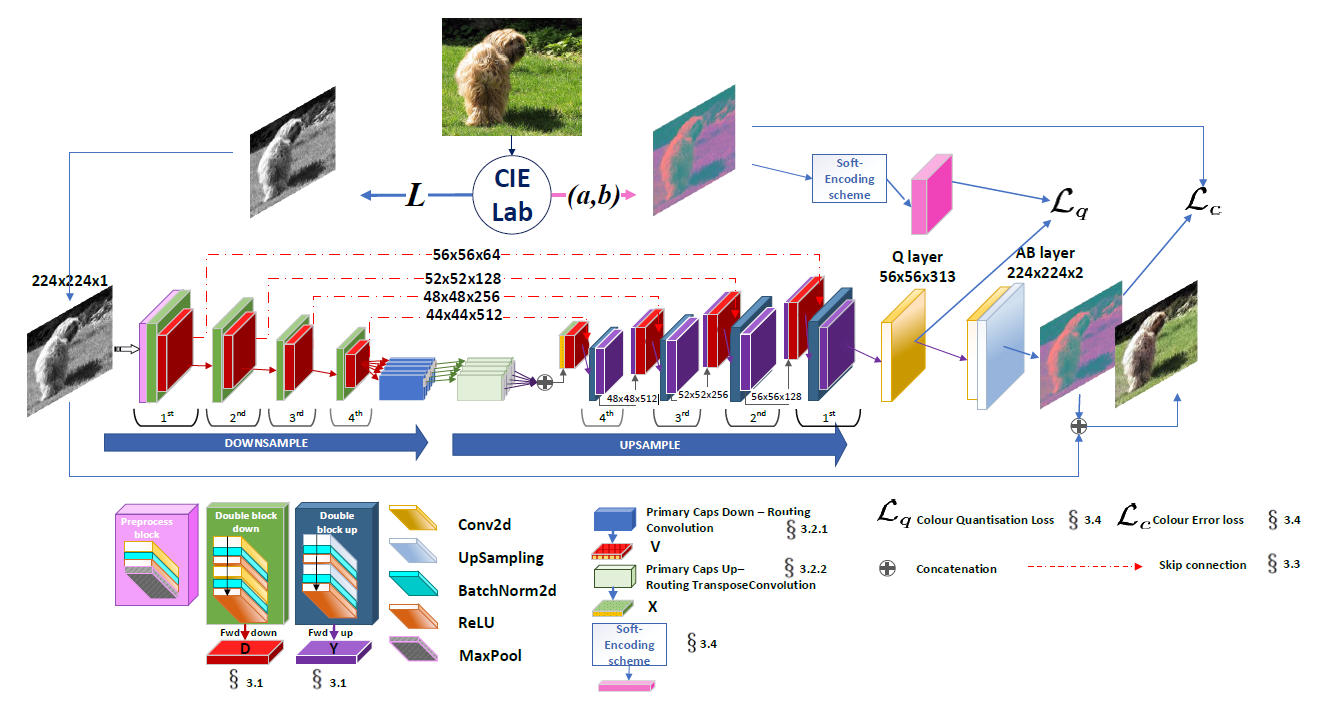}
\end{center}
   \caption{Our proposed \myalgoname architecture takes the $L$ channel of a CIELab image and processes it by a set of \textit{double block down} layers detecting image-level features. Then \textit{primary caps down} layers extracts the entity-level features.
   Such disentangled representations are then considered in a quantisation layer that allows us to learn a colour distribution, hence to generate plausible $(a,b)$ colours channels.}
\label{fig:UCapsNet}
\end{figure*}
\section{Related works}

Existing image colourisation works are either based on scribble and example images, or rely on learning processes.
\textbf{Based on Scribbles and Example Images.}
The multi-modal nature of image colourisation  problem, was initially tackled with local hints provided by humans \cite{levin2004colorization,huang2005adaptive,qu2006manga,luan2007natural,yatziv2006fast,zhang2017real}.
These approaches propagate similar colourisations over specific areas, recognised with low-level similarity metrics. In~\cite{levin2004colorization}, pixels with similar luminescence should have a similar colour, given by the scribble. This method suffers of colours overcoming object's edges, later limited by means of edge detection~\cite{huang2005adaptive} and texture similarity~\cite{qu2006manga,luan2007natural}. An up-grade was proposed in~\cite{yatziv2006fast} by relaxing constraints on the position of scribbles. In~\cite{zhang2017real}, scribbles and human hints were jointly exploited to limit the manual effort. Other works explore the ``example image'' technique where colours are transferred from a reference (example) image to a greyscale image. The reference images are specified by users or searched on internet~\cite{welch2002transfer,charpiat2008automatic,gupta2012image,chia2011semantic,ironi2005colorization,he2018deep}. Information is transferred between reference and input images on the basis of low-level similarity metrics measured at pixel level~\cite{welch2002transfer}, semantic segments level~\cite{charpiat2008automatic} or super-pixel level~\cite{gupta2012image, chia2011semantic}.
A combination of scribbles and example images was proposed in~\cite{ironi2005colorization,he2018deep}.

Even if the achieved results are interesting for some specific tasks, the required human effort is too intensive and generally not affordable when the colourisation objective considers large scale problems.
Our approach is fully automated, thus suitable for large scale colourisation tasks.

\textbf{Based on Learning Processes.}
The colourisation task through machine learning has received great attention in the recent past~\cite{iizuka2016let,zhao2018pixel,zhang2016colorful,larsson2016learning,mouzon2019joint,isola2017image,vitoria2020chromagan,deshpande2015learning,deshpande2017learning,cheng2015deep,ozbulak2019image,guadarrama2017pixcolor}.
In~\cite{iizuka2016let,zhao2018pixel} semantic labels are exploited to capture entity-level features that, together with image-level features, improve the colourisation.
In~\cite{zhang2016colorful}, semantic interpretability is obtained by a cross-channel encoding scheme, later exploited in~\cite{mouzon2019joint} with a pre-trained classification model.
Generative models~\cite{isola2017image, nazeri2018image,vitoria2020chromagan} were also exploited together with pre-trained classifiers. 
Other approaches~\cite{zhang2016colorful,larsson2016learning,zhao2018pixel} focused on multi-task models and single pixel significance. 

Recently, more focus was devoted to capture entity-level features by embracing the capsules concept~\cite{ozbulak2019image} or by exploiting pre-trained object detection models~\cite{Su2020CVPR}

Such approaches either separately considered image-level and entity-level features or hinge on models pre-trained through supervised approaches.
In contrast, we jointly capture the two representations and introduce a model that is trained with self-supervised signals only. 

The works in~\cite{ozbulak2019image} and~\cite{Su2020CVPR} are the closest to our approach.
Differently from these, we introduce a single architecture that
(i) captures entity-level features through capsules without using any prior knowledge to detect and extract object information;
(ii) obtains the spatial information discarded by capsules through image-level features from CNNs;
(iii) tackles the emerging problem of figure-ground separation by enforcing collaboration between the disentangled image-level and entity-level features with the introduction of skip connections among layers computing such two representations.
We also introduce (iv) a combination of a colour quantisation loss~\cite{zhang2016colorful} to learn a distribution of colours and a colour error loss that allows us to deal with multiple plausible colourisations of a same/similar object.

 ==============================================================================================================================

\section{Proposed Approach}
Our goal is to generate a plausible colourisation for a grayscale input image.
Towards such a goal, our architecture, shown in \figurename~\ref{fig:UCapsNet}, starting from the CIELab lightness channel $L\in \mathbb{R}^{H\times W\times 1}$ learns to predict the corresponding colour channels $(a,b)\in\mathbb{R}^{H\times W\times 2}$.
These represent the self-supervised training signal.

\textit{The downsample phase} is responsible of learning disentangled image-level and entity-level representations.
The former is obtained through a preprocessing block followed by a set of consecutive \textit{double block down} operators which include convolutional layers.
The latter is computed by the \textit{primary capsule down} operator taking the image-level features to encode the entity-level ones into capsules.

\textit{The upsample phase} leverages the disentangled image-level and entity-level representations to generate a plausible image colourisation. 
The \textit{primary capsules up} operator decodes the input capsules to transform them into spatial features carrying information about object entities.
Such features are combined with image-level features obtained through skip connections, then processed by a set of consecutive \textit{double block up} operators.

The model learns a colour distribution over pixels ($\mathbf{Q}$ layer) that is later exploited to predict the colour channels ($\mathbf{AB}$ layer).

\subsection{Image-Level Features}
\label{sec:DBConv}
Convolutional layers extract a feature map indicating the location and strength of a detected feature in an image.
These define the image-level representation carrying the context information that is important to perform a suitable colourisation.
We obtain such a representation through the introduction of a preprocessing block, followed by the \textit{double block down} (\doubleblockdown).
The former is composed of a \myconv--\mybn--\myrelu--\mymaxpool sequence.
To design the latter, we followed a common practice~\cite{hadji2018we} by which image-level features can be obtained through a hierarchical structure of repeated convolutional, normalisation, and non-linearity layers.
Thus, we let a \doubleblockdown be composed of two consecutive sequences of \myconv--\mybn--\myrelu layers. 
Four \doubleblockdown are stacked in the downsample phase with $\mathbf{D}_i$ being the output of $\doubleblockdown_i$, for $i=1,\cdots,4$.

\subsection{Entity-Level Features}

While image-level features are important to capture information about the context of an image, the colourisation process also hinges on the entities present in the scene.
To capture entity-level features, we exploit the concepts behind capsules~\cite{sabour2017dynamic},~\ie, groups of neurons generating vectors that indicate the probability of existence of object entities.

\subsubsection{Primary Caps Down (\primarycapsdown)}
The \primarycapsdown operator is composed of two capsule layers.
The first layer of capsules computes $\mathbf{U} = [ \myflatten(\myconv_1(\mathbf{D}_4))^T, \cdots, \myflatten(\myconv_k(\mathbf{D}_4))^T]$.
With each column of $\mathbf{U}$ being the capsule output $\mathbf{u}_i\in\mathbb{R}^k$.
To identify entity-level features, a weight matrix $\mathbf{W}_{ij} \in \mathbb{R}^{k\times \hat{k}}$ is introduced to obtain the second layer capsules $\hat{\mathbf{u}}_{j|i}=\textbf{W}_{ij}\mathbf{u}_i$, later grouped through the ``routing by agreement'' mechanism in Algorithm~\ref{algo:routing}.
\begin{algorithm}[t]
\SetAlgoLined
\SetKwInput{KwInput}{Input}
\SetKwInput{KwOutput}{Output}
\KwInput{Weighted capsule features $\hat{\mathbf{u}}_{j|i}$}
\KwOutput{Entity-level feature vector: $\mathbf{v}_j$}
 \text{for all capsule} $i$ \text{in the first layer and capsule} $j$ \text{in the second layer}: $b_{ij} \leftarrow 0$ \\
 \For{\text{routing iterations}}
 {
  \text{for each capsule} $i$ \text{in the first layer}: $\mathbf{c}_i \leftarrow \text{softmax}(\mathbf{b}_i)$\;
  \text{for each capsule} $j$ \text{in the second layer}: \ $\mathbf{s}_j \leftarrow \sum_{i}c_{ij}\hat{\mathbf{u}}_{j|i}$\;
  \ \ $\mathbf{v}_j \leftarrow \text{squash}(\mathbf{s}_j)$; \qquad $\triangleright \text{squash}$ as in~\cite{sabour2017dynamic} \\
  for each capsule $i$ in the first layer and capsule $j$ in second layer: \ $b_{ij} \leftarrow b_{ij} + \hat{\mathbf{u}}_{j|i} \mathbf{v}_j$
 }
 \caption{Routing by Agreement~\cite{sabour2017dynamic}}
 \label{algo:routing}
\end{algorithm}
At each iteration of the routine, $\hat{\mathbf{u}}_{j|i}$'s are grouped in agreement with the coupling coefficient $\mathbf{c}_i$ to identify clusters of features,~\ie, the entity-level feature vector $\mathbf{v}_j \in \mathbb{R}^{\hat{k}}$.
This carries information about how strong the capsules agree on the presence of an entity.

\subsubsection{Primary Caps Up (\primarycapsup)}
The \primarycapsdown capsule outputs $\mathbf{v}_j$'s contain entity-level features but lack details about their spatial displacement. Since this information is fundamental for the colourisation task, we introduce a mechanism that inverts the \primarycapsdown procedure to reconstruct the spatial information.
We introduce a weight matrix $\mathbf{W}^{r}_{ji} \in \mathbb{R}^{\hat{k}\times k}$ connecting each \primarycapsdown output to the \primarycapsup capsules.
This computes $\mathbf{u}^{r}_{i} = \mathbf{W}^r_{ji}\mathbf{v}_j$, that are then stacked to obtain $\mathbf{U}^r$ --with the same size of $\mathbf{U}$.
The resulting $k$ rows of $\mathbf{U}^r$ are then reshaped, processed by $k$ independent \myTconv operators, then concatenated to obtain an output matrix $\mathbf{X}$ having the same dimensionality of $\mathbf{D}_4$.
Thus, $\mathbf{X}$ contains spatial information generated through the entity-level features obtained through capsules.

\subsection{Image-Level and Entity-Level Collaboration}
Through the introduction of the capsule layers, we are able to extract entity-level features. However, generating the colourisation output requires a precise answer at pixel level, which is likely not to be directly achievable through the \primarycapsup layer alone (see \S~\ref{sec:ablation}).
Thus, to promote such an output, we enforce collaboration between image-level and entity-level features by introducing a mechanism to project the image-level features learn at different stages of the downsample phase into the pixel space~\cite{ronneberger2015u}.
This is achieved by introducing skip connections from the downsample phase to the upsample one. The upsample phase is composed of four \textit{double block up} ($\doubleblockup$) operators designed following the same considerations adopted for the \doubleblockdown, hence composed of two consecutive sequences of \myupsample-\mybn-\myrelu layers.
As shown in~\figurename~\ref{fig:UCapsNet}, $\doubleblockup_i$, with $i=4,\cdots,1$ receives as input the concatenation of $\mathbf{D}_i$ and the output of the preceding $\doubleblockup_{i+1}$, denoted as $\mathbf{Y}_{i+1}$.
$\doubleblockup_4$ processes the concatenation between the de-routed entity-features in $\mathbf{X}$ and $\mathbf{D}_4$.
The up-sampling operations in each \doubleblockup allow us to reconstruct the $(a,b)$ channels having the same size of the input image.
The skip connections enforce exploitation of higher resolution features that can be missing due to the sparsity of the up-sampling operations.

\begin{figure*}[t]
\begin{center}
\includegraphics[width=\textwidth,height=0.9\textheight,keepaspectratio]{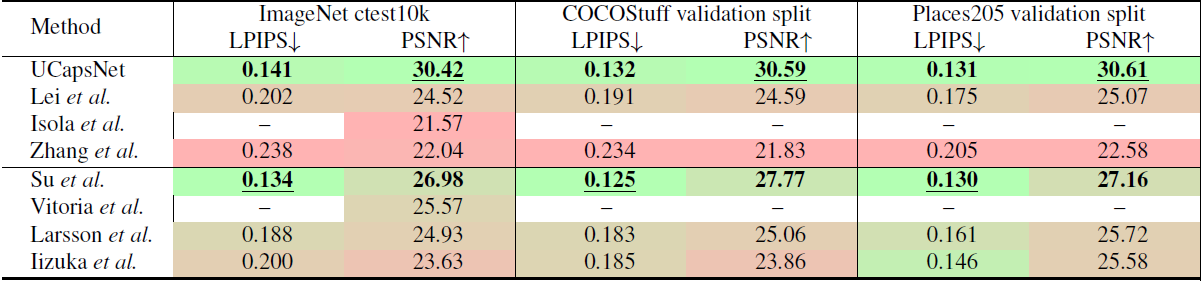}
\end{center}
   \caption{Quantitative comparison.
    Top rows show the results achieved by approaches that do not consider labels for training nor models pre-trained with supervision (\ie, unsupervised/self-supervised methods).
    Bottom rows show the performance obtained by methods that need semantic labels for training or require models pre-trained with supervision  (\ie, supervised methods). 
    Columns follow a red-to-green colour-coded representation: the better the performance the greener the table cell. 
    Best results for each of the two groups are in bold. Best overall results are also underlined. (\textit{Best viewed in colours.})}
\label{tab:quantitativeCompare}
\end{figure*}

\subsection{Objective Function}
\label{sec:objectiveFunction}
A same/similar object can have different colourisations.
To allow the model capture such a multi-modality of appearance of entities, we propose a \textbf{Q} layer to learn a distribution over quantised pixel colours.
However, learning such a distribution is not the final objective of the colourisation process.
We ultimately want the model to generate the chrominance $(a,b)$ colours for the $L$ input.
Towards such a goal, we exploit the learned distribution to produce a plausible colourisation.
This is achieved through the \textbf{AB} layer learning a mapping from the quantised space to the chrominance one by means of the colour error loss.

\subsubsubsection{Colour Quantisation Loss}
To generate a plausible colourisation, we want to learn a distribution over per-pixel colours.
Towards such an objective, inspired by~\cite{zhang2016colorful}, we quantised the $(a,b)$ space into bins with grid size 10.
We then kept only the $Q=313$ values which are in-gamut.
These denote the distinct classes a pixel can belong to.
 Starting from the input channel $L$, our model learns to generate a distribution over such classes.
This is achieved by introducing the \textbf{Q} layer composed of $\myupsample$ and $1\times 1$-\myconv layers, processing the $\mathbf{Y}_1$ feature map to predict the colour distribution $\mathbf{\hat{Z}}\in\mathbb{R}^{H\times W\times Q}$.
This is used to compute the quantisation loss
\begin{equation}
    \mathcal{L}_{q} = -\sum_{h,w}v(\mathbf{Z}_{h,w}) \sum_{q}\mathbf{Z}_{h,w,q}log(\mathbf{\hat{Z}}_{h,w,q})
    \label{eq:lq}
\end{equation}
where $\mathbf{Z}_{h,w,q}$ is the ground-truth colour distribution for the $(h,w)$ pixel obtained through a soft-encoding scheme and $v(\cdot)$ re-weights the loss for each pixel based on pixel colour rarity.
We have considered the soft-encoding and the $v(\cdot)$ values introduced~\cite{zhang2016colorful}.

\subsubsubsection{Colour Error loss}
Our final objective is to generate the $(a,b)$ chrominance channels.
To achieve this, the \textbf{AB} layer takes $\mathbf{\hat{Z}}$ as input and processes it with a $1\times 1$-\myconv layer reducing the $Q$ feature maps to 2.
These represent the predicted $(\hat{a},\hat{b}) \in \mathbb{R}^{H\times W \times 2}$ channels obtained by minimising their difference with the real chrominance ones $(a,b)$ as:

\begin{equation}
\mathcal{L}_{c} = ||\hat{a}-a||^2_2 + ||\hat{b}-b||^2_2.
\label{eq:lc}
\end{equation}

\subsubsubsection{Combined Loss}
We optimise our model for $\mathcal{L}_q+\mathcal{L}_c$.
This allows us to generate a plausible colourisation for a same/similar object by exploiting the quantised distribution (learned through $\mathcal{L}_q$) while avoiding the weakness of using $\mathcal{L}_{c}$ alone, which produces desaturated colours~\cite{zhang2016colorful}.

\section{Experiments}
\label{sec:experiments}
To validate our approach, we present extensive experimental results on three benchmark datasets using different evaluation metrics (\S~\ref{sec:settings}).
Precisely, in \S~\ref{sec:soacomparison}, we validate and compare our colourisation performance with recent and relevant works~\cite{Su2020CVPR}, also through a large scale user-study.
In \S~\ref{sec:ablation}, we evaluate the key components of our approach. 
In \S~\ref{sec:selfsupervision}, we test colourisation as a method for self-supervised representation learning.
Finally, in \S~\ref{sec:legacy}, we show qualitative examples on legacy black and white images.

\subsection{Settings}
\label{sec:settings}
\subsubsubsection{Datasets}
To assess the performance of our approach, we have considered three benchmark datasets coming with different features and colourisation challenges.

\textbf{\textit{ImageNet}\cite{russakovsky2015imagenet}} is widely used as colourisation benchmark.
The 1.3M training images (with no labels) were considered for model training for all the following experiments.
The ctest10k~\cite{larsson2016learning} samples have been used for evaluation.

\textbf{\textit{COCOStuff}\cite{caesar2018coco}} contains a wide variety of natural scenes with multiple objects present in the 118k images.
We used the provided validation split containing 5000 images.

\textbf{\textit{Places205}~\cite{zhou2014learning}} is a scene-centric dataset containing samples 205 different categories.
We considered the 20500 validation images.  

Note that, COCOStuff and Places205 have been used only for evaluating the colourisation transferability.
We do not use the corresponding training sets to learn the model parameters.

\subsubsubsection{Evaluation metrics}
\label{sec:metrics}
To assess the colourisation quality, we followed the experimental protocol proposed in~\cite{Lei_2019_CVPR} and considered the Peak Signal to Noise Ratio (PSNR) and the the Learned Perceptual Image Patch Similarity (LPIPS)~\cite{zhang2018perceptual} (version 0.1 with VGG backbone).


\subsubsubsection{Implementation Details\protect\footnote{Code will be publicly released}}
We train our network for 20 epochs with a batch size of 32 on the 1.3M ImageNet training samples (with no labels) resized to $224\times 224$.  Each image is first projected into the CIELab colourspace, then the resulting $L$ channel is used as the input. The $(a,b)$ channels are considered to compute the $\mathcal{L}_{q}$ (after quantisation) and $\mathcal{L}_{c}$ losses. We used the Adam optimiser with a learning rate of $2\times 10^{-5}$.
The precise details of the whole \myalgoname architecture are in the supplementary material.
Using the PyTorch framework, a single epoch takes about 10 hours on an NVidia Titan RTX.

\begin{figure*}[h!]
\begin{center}
\includegraphics[width=\textwidth,height=0.9\textheight,keepaspectratio]{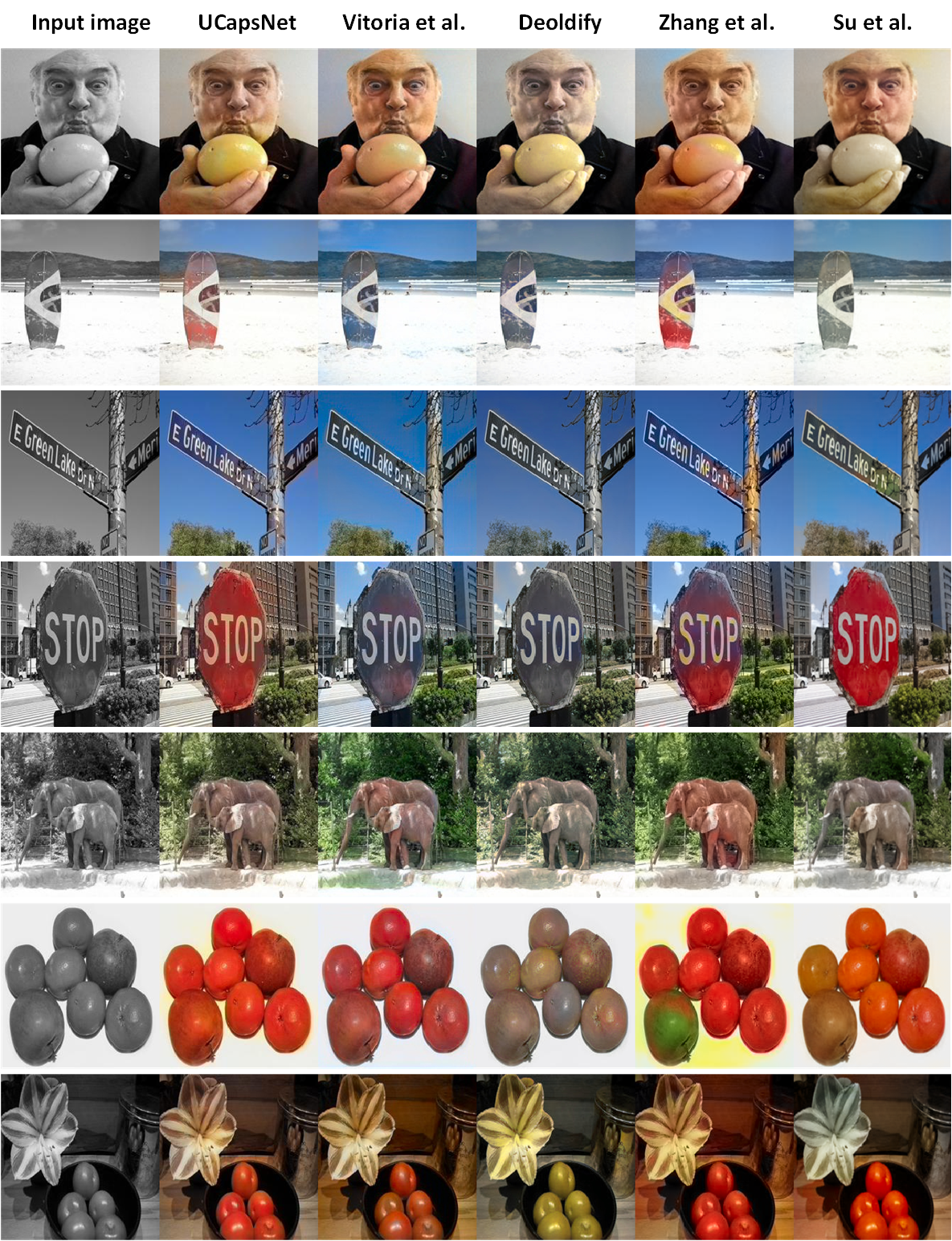}
\end{center}
   \caption{Comparison between state-of-the-art automatic colourisation methods. Our method generates vivid colours that are well defined inside the contours of the entities and contain no splotches. Images are plausible and pleasing for different complex scenes with multiple object instances.
   Results obtained on COCOStuff dataset. }
\label{fig:ComparisonQuality}
\end{figure*}
\begin{figure*}
\begin{center}
\includegraphics[width=\textwidth]{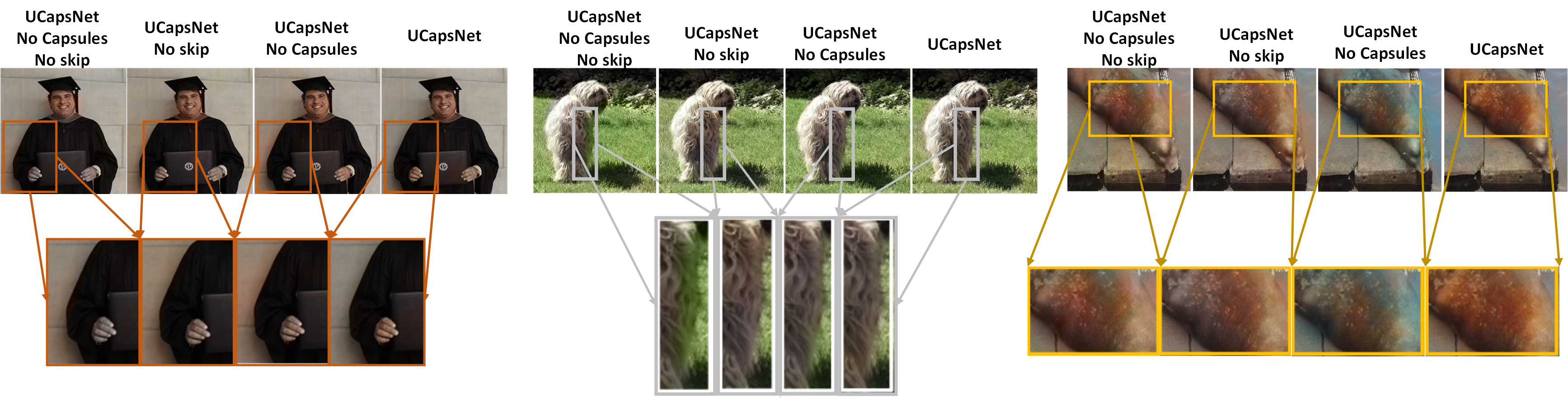}
\end{center}
   \caption{
   Analysis of the importance of image-level and entity-level features, and their collaboration in UCapsNet on three image samples from COCOStuff.
   The first row compares the predicted colourisations of a baseline model where capsules and skip connections are not considered (UCapsNet No Capsules No Skip) with three variants were we included (i) capsules (\myalgoname No Skip), (ii) skip connections (\myalgoname No Capsules), and (iii) both together (\myalgoname).
   In the second row, an expanded view showing detailed differences on the predicted images. These demonstrate the advantage of exploiting both disentangled representations, then enforcing their collaborations to produce a plausible colourisation.
   }
   \label{fig:PicInPic}
\end{figure*}

\subsection{State-of-the-art Comparisons}
\label{sec:soacomparison}
\subsubsection{Quantitative Performance}
\subsubsubsection{Standard Evaluation Metrics}
Table in Figure~\ref{tab:quantitativeCompare} shows the current leaderboard on the three considered datasets.
Results show that our approach outperforms all existing solutions in terms of PSNR with a significant margin over methods exploiting labelled data,~\ie, we increase the PSNR of~\cite{Su2020CVPR} by more than $10\%$ on the three datasets on average.
LPIPS performance significantly improve with respect to existing methods that do not using semantic labels for training or exploit pre-trained models with supervision (\eg, $0.141$ vs $0.202$ obtained by~\cite{Lei_2019_CVPR}).
Concerning approaches exploiting labelled data, our method achieves very similar LPIPS performance to~\cite{Su2020CVPR} which hinges on a pre-trained model for object detection.
Note that for Places205 the performance difference with~\cite{Su2020CVPR} is only of $0.1\%$.
Such results demonstrate that our approach is able to capture relevant image and entity features to achieve competitive results, without the need of explicit information (\eg, labelled data or dedicated object detection models).

\subsubsubsection{User Study}
We conducted a user study to quantify the perceptual realism of the colourisation results obtained with our method in comparison with Vitoria~\etal~\cite{vitoria2020chromagan}, DeOldify~\cite{Jantic}, Zhang~\etal~\cite{zhang2016colorful}, and Su~\etal~\cite{Su2020CVPR}.
We randomly selected 200 images from the COCOStuff validation dataset.
We show to each participant 20 image pairs composed of our generated image and an image generated by one of the above methods (randomly) and we asked for preference.
We collected a total of 3600 votes from 180 participants.
Results show that our method is preferred over Vitoria~\etal~\cite{vitoria2020chromagan}~(53\% vs 47\%), DeOldify~\cite{Jantic}~(64\% vs 36\%), Zhang~\etal~\cite{zhang2016colorful}~(54\% vs 46\%), and Su~\etal~\cite{Su2020CVPR}~(54\% vs 46\%).
From a qualitative analysis, we noted that images with saturated/vivid colours are preferred even in presence of colour splotches or implausible colours.

\subsubsection{Qualitative Performance}
\figurename~\ref{fig:ComparisonQuality} compares the colourisation results of our approach and competing solutions on the COCOStuff dataset. 
In general, we notice that our method provides a consistent object/background separation also reducing the colour blurring on contours thus generating more detailed outputs.
Colours look vivid in all the images, and this is more evident when our results are compared with Deoldify~\cite{Jantic}, Vitoria~\etal~\cite{vitoria2020chromagan} and Su~\etal~\cite{Su2020CVPR} (\eg, examples at rows 1, 2, 5, and 7).
With respect to Zhang~\etal~\cite{zhang2016colorful} (fifth column), our generated colours are better defined inside the contours of the entities and contain no splotches (\eg, examples at rows 3, 4, 6).
Finally, it is worth noting UCapsNet provides specific colours entangled with the nature of the entities in the image (\eg, examples at rows 1, 2, 4).

\subsection{Ablation Study}
\label{sec:ablation}
We validate the main design choices of our model by analysing the role of the key architecture components, then we evaluate the effects of the considered loss functions.

\begin{figure*}
    \centering
    \includegraphics[width=\textwidth]{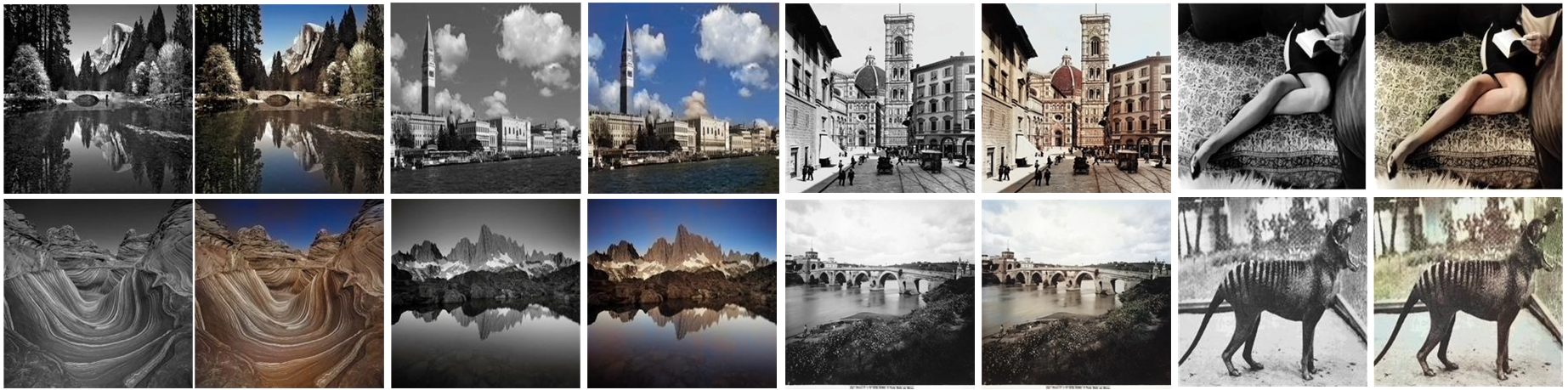}
    \caption{Applying UCapsNet to black and white photographs from Henri Cartier Bresson, Ansel Adams, Alinari. Historical archives.}
    \label{fig:LegalyBW}
    \vspace{-1.2em}
\end{figure*}

\subsubsubsection{\myalgoname Variants}
To assess the importance of image-level features, entity-level features and their collaboration for the colourisation objective, we start from the model baseline (\myalgoname No Capsules, No Skip) and progressively included our contributions.
In~\figurename~\ref{fig:PicInPic}, we report on the results obtained considering (i) capsules (\myalgoname No Skip), (ii) skip connections (\myalgoname No Capsules) and (iii) both (\myalgoname).
The baseline model (\myalgoname No capsules No skip) does not respect object boundaries,~\eg, the fur of the dog is partially green as the grass, and generates inconsistent entity colourisations, like the hands of the man which are grey.
By adding entity-level features (\myalgoname No Skip), object are recognised and their boundaries respected. 
No splotches are also present in the results.
When skip-connections are included and capsules rejected (\myalgoname No Capsules), the colourisation overflows the object boundaries, thus demonstrating that capsules carry important entity-level features,~\eg, the seal image, where the animal shape is not recognised.
Such results substantiate the importance of disentangling the image-level and entity-level features then promoting their collaboration to generate a well defined colourisation,~\ie, \myalgoname results.

\begin{figure}[t]
\begin{center}
\includegraphics[width=0.8\linewidth]{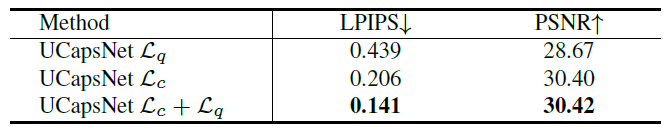}
\end{center}
   \caption{Analysis of the importance of the considered losses. Results are computed for UCapsNet optmised by considering the loss functions defined in \S~\ref{sec:objectiveFunction}. Best results are in bold.\vspace{-0.5em}.}
\label{tab:ablation}
\end{figure}

\begin{figure}[t]
\begin{center}
\includegraphics[width=\linewidth]{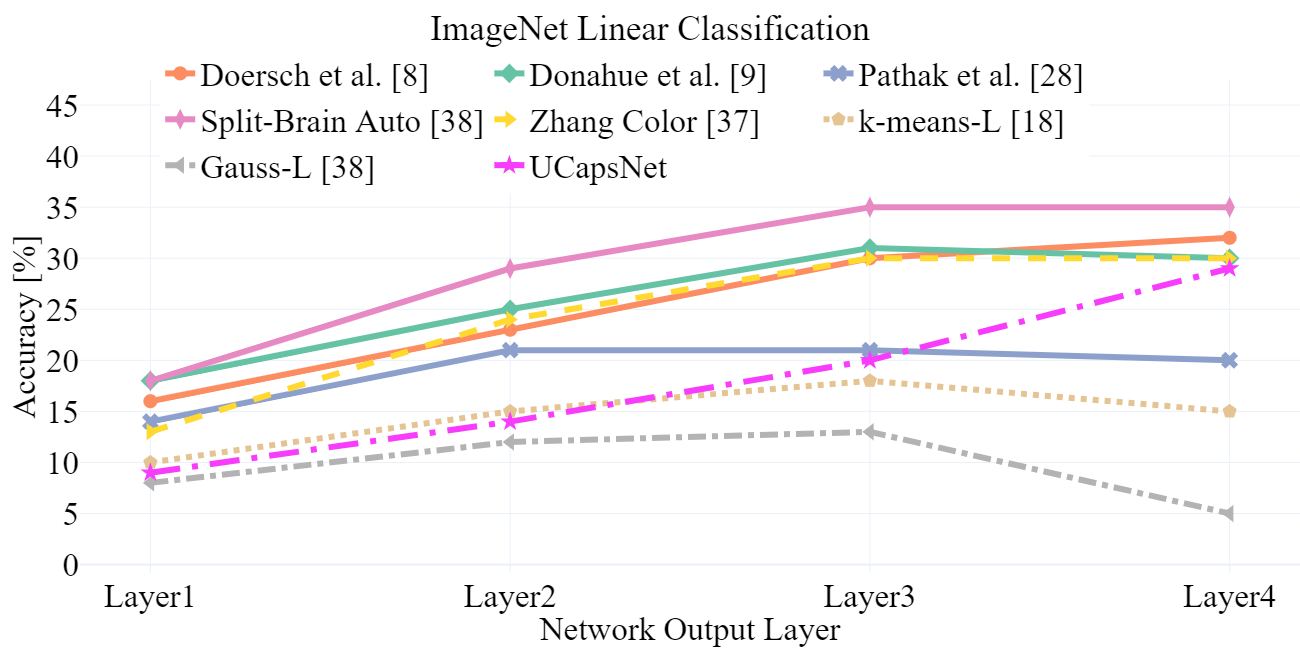}
\end{center}
   \caption{Task Generalisation on ImageNet. We freeze each downsampling layer and evaluate the features extracted at different depths by stacking a classification layer on top of those.
   Methods working on a greyscale input are shown with dotted lines while approaches considering an RGB input are with solid lines.}
\label{fig:Linear}
\vspace{-1.4em}
\end{figure}

\subsubsubsection{Losses}
To evaluate the role of the loss functions composing our optimisation objective, we have trained our model separately considering $\mathcal{L}_{q}$ and $\mathcal{L}_{c}$.
LPIPS and PSNR performance in Table in Figure~\ref{tab:ablation} show that the combination of the two losses improves the results achieved by exploiting either $\mathcal{L}_{c}$ or $\mathcal{L}_{q}$.
Such a result substantiates the importance of jointly learning a colour distribution as well as a mapping to the chroma channels to generate a plausible colourisation able to handle the  multi-modality of appearance of entities.


\vspace{-0.3em}
\subsection{Self-Supervised Colourisation as Pretext}
\label{sec:selfsupervision}
\vspace{-0.2em}
In~\figurename~\ref{fig:Linear}, we evaluate the trained model for representation learning. For a fair comparison, we followed the protocol in~\cite{zhang2016colorful} and evaluated the generalisation capability of the convolutional features learned in the downsampling phase, hence excluded the entity-level representations.
At the output of each \doubleblockdown, we stacked a $\mymaxpool$ layer, with equal kernel and stride sizes such that the generated feature dimensionality is below 10k~\cite{zhang2016colorful}.
Then, we added a linear classifier for the ImageNet classes that is trained for 100 epochs. In~\figurename~\ref{fig:Linear}, we compare with methods working on RGB inputs~\cite{doersch2015unsupervised,donahue2016adversarial,pathak2016context,zhang2017split}, with models trained on grayscale images~\cite{zhang2016colorful} (like we do) and with models initialized with random/Gaussian weights or using the k-means scheme~\cite{krahenbuhl2015data}.
Performance obtained with \textit{layer1} output (\ie, $\doubleblockdown_4$ in our architecture), is in line with the methods based on grayscale input.
With increasing depths,~\ie, at \textit{layer3} and \textit{layer4} our method reaches competitive results. The classification accuracy obtained with the \textit{layer4} features (\ie, $29\%$) is in line with existing methods working on an RGB input.
This shows that, despite the input handicap, representations obtained considering our colourisation method as a pretext task carry relevant information that can be used to discriminate among different semantic classes. 


\subsection{Colourising Legacy Black and White Photos}
\label{sec:legacy}
Our model is trained using generated grayscale images by removing the $(a,b)$ channels from coloured photos.
In~\figurename~\ref{fig:LegalyBW}, we show that \myalgoname is able to produce realistic colourisations on real legacy black and white photos, even though the low-level statistics of the such photos are different from the modern-day images used for training.

\vspace{-0.3em}
\section{Conclusion}
\vspace{-0.2em}
Image colourisation is an instance of a difficult pixel prediction problem in computer vision. We have demonstrated that disentangling image-level (through CNN) and entity-level (via capsules layers) features, then enforcing collaboration among them, produces results indistinguishable from real colour photos.
Without any prior knowledge, we are able to colourise a greyscale picture respecting the relevant image details and entities differences.
Through extensive experiments, we demonstrated that our method outperforms existing colourisation approaches methods that hinge on labelled training data.
We have also shown that our model works very well as a pretext task for representation learning, performing strongly compared to other self-supervised pre-training methods.

\bibliographystyle{ieee_fullname}
\bibliography{egbib}

\end{document}